# Technical Analysis on Financial Forecasting

S.Gopal Krishna Patro[1], Pragyan Parimita Sahoo[2], Ipsita Panda[3], Kishore Kumar Sahu[4]

[1,2,3,4] *Department of CSE & IT, VSSUT, Burla, Odisha, India*

sgkpatro2008@gmail.com, pragyansahoo2@gmail.com, pipsita47@gmail.com, itkishore2000@gmail.com

*Abstract*— Financial forecasting is an estimation of future financial outcomes for a company, industry, country using historical internal accounting and sales data. We may predict the future outcome of BSE_SENSEX practically by some soft computing techniques and can also optimized using PSO (Particle Swarm Optimization), EA (Evolutionary Algorithm) or DEA (Differential Evolutionary Algorithm) etc. PSO is a biologically inspired computational search & optimization method developed in 1995 by Dr. Eberhart and Dr. Kennedy based on the social behaviors of fish schooling or birds flocking. PSO is a promising method to train Artificial Neural Network (ANN). It is easy to implement then Genetic Algorithm except few parameters are adjusted. PSO is a random & pattern search technique based on populating of particle. In PSO, the particles are having some position and velocity in the search space. Two terms are used in PSO one is Local Best and another one is Global Best. To optimize problems that are like Irregular, Noisy, Change over time, Static etc. PSO uses a classic optimization method such as Gradient Decent & Quasi-Newton Methods. The observation and review of few related studies in the last few years, focusing on function of PSO, modification of PSO and operation that have implemented using PSO like function optimization, ANN Training & Fuzzy Control etc. Differential Evolution is an efficient EA technique for optimization of numerical problems, financial problems etc. PSO technique is introduced due to the swarming behavior of animals which is the collective behavior of similar size that aggregates together.

## I. INTRODUCTION

As we have studied that the forecasting is the important role play in the market or big challenge in the current market, forecasting is the term which can be applied in the field of Banking, Finance, Company or Industry for their future outcome prediction with respect to the past historical analysis. So there are so many concepts the industry or company are using for their forecasting purpose. But we have chosen a best concept for our forecasting purpose namely Soft Computing and mainly this concept also using by the industry or company from last hundreds of years back. And the details soft computing technique that we are using for our research purpose will be discuss in Section II, namely Neural Network & optimization technique for getting better result after optimization is Particle Swarm Optimization and many more.

   We describe the basic approach to forecast the financial data [1][10] using Dataset (as a historical data for forecasting of future outcome), Neural Network (as a learning paradigm), and Particle Swarm Optimization (for optimization purpose). We will also describe about the different technique for choosing the Inputs, Outputs and calculation of Errors by using transfer function.

## II. METHODOLOGY

The technical details of forecasting the sensex data is to be carried out as follows:

*Step*-**1.** Take the Historical Data [11].

| BSE_SENSEX DATA_MONTHLY | | | | |
|---|---|---|---|---|
| **Month** | **Open** | **High** | **Low** | **Close** |
| Jun11 | 19859.22 | 19860.19 | 18467.16 | 19395.81 |
| Jul11 | 19352.48 | 20351.06 | 19126.82 | 19345.7 |
| Aug11 | 19443.29 | 19569.2 | 17448.71 | 18619.72 |
| Sept11 | 18691.83 | 20739.69 | 18166.17 | 19379.77 |
| Oct11 | 19452.05 | 21205.44 | 19264.72 | 21164.52 |
| Nov11 | 21158.81 | 21321.53 | 20137.67 | 20791.93 |
| . | | | | |
| . | | | | |
| . | | | | |
| Dec14 | 20771.27 | 21483.74 | 20568.7 | 21170.68 |

Table 1 Sample of BSE_SENSEX Data

*Step*-**2.** Convert unstructured data into structured data or Normalize data [12].

   Normalization can be done by any one of the following technique:

$$X = \frac{2Y - (Max + Min)}{(Max - Min)}$$

(Or)

$$X = \frac{(HIgh - Low)}{(Max - Min)} \times (Y - Min)$$

(Or)

$$X = \frac{Low + (Y - Min)(High - Low)}{(Max - Min)}$$

where,  $Y$ is the current position
 $Max$ is the Maximum value of column
 $Min$ is the Minimum value of column
 $High$ is the higher value of chosen range
 $Low$ is the lower value of chosen range

***Step*-3.** Input the normalized data into the Neural Network.

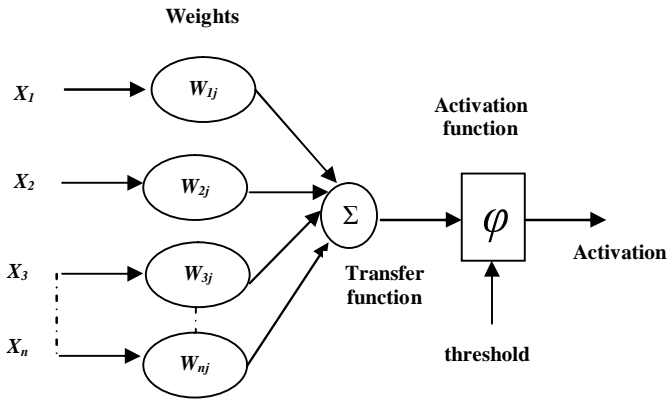

Fig.1 Neural Network

***Step*-4.** Selection of best optimization technique like PSO, ACO for selection of neural Network [2].

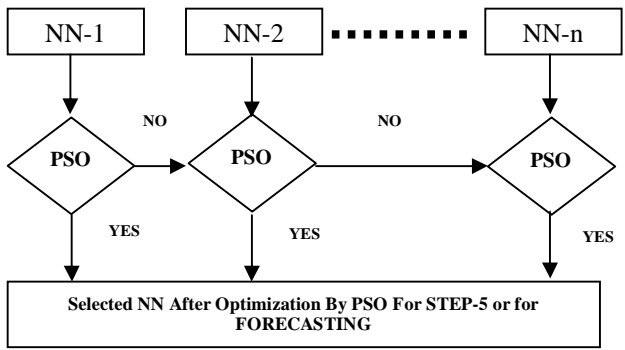

Fig.2 PSO for NN Selection

***Step*-5.** Forecast the future outcomes by using best one Neural Network which is selected through optimization technique in Step-4.

## III. FINANCIAL FORECASTING

There are so many papers are there where they are dealing with financial market forecasting. Many of them are using soft computing techniques like Neural Network [12], Neuro-Fuzzy Inference System [3], Genetic Algorithm [4], Particle Swarm Optimization [5], Ant Colony Optimization, Multi-Objective Particle Swarm Optimization and many more. But as per human being knowledge is impossible to correctly predict the exact market value. Therefore, researcher or author has used complicated method or so many methods to minimize the forecasting error. Instead of finding the exact one to find the predicted one value we have proposed here a method or technique which will minimize the error and predict the future outcome of the finance market.

Some real time forecasting is used for the forecasting purpose are as follows:

- Bayesian method
- Reference class forecasting
- Proforma Financial Statement
- Budget expense method

Out of these four methods Bayesian method is the best one for the forecasting for our research work. Because this method especially depends upon the historical data or scenario before it becomes predict the future outcome.

$$P(A|B) = \frac{P(A \cap B)}{P(B)} = \frac{P(A) * P(B|A)}{P(B)}$$

where,  P($A$) is the probability of $A$ occurring
 P($B$) is the probability of $B$ occurring
 P($A|B$) is the conditional probability of $A$ given that B occurs.
 P($B|A$) is the conditional probability of $B$ given that A occurs.

## IV. NEURAL NETWORK

The Neural Network concept has been first introduced by Warren McCulloch & Walter pitts around 1943 based on the mathematics and algorithm [12]. And the Learning process has been introduced around 1948 by Donald Hebb.

### A. Feed-foreward networks

Feed foreward networks are refers to the straight foreward networks which is associated with input to the output only. There is no feed back path to the same layer that's output of any layer does not effect to the same layer.

## B. Feedback networks

Feedback networks are also called as recurrent networks. Feedback networks are very much powerful network and also complicated type of networks. Feedback networks are dynamic type of networks, where the state is continuously changing until they reach in a particular point.

## C. Network Layers

Network layer consist of three layer namely Input layer (This layer can take the row information into the network), Hidden Layer (This layer identifies the activities of input layer and weight associated with the input layer and hidden layer) & Output layer (depends upon the activity of hidden layer and weight associated with hidden layer and output layer).

## D. Learning process

Learning is the process for determination of the weight. This can be applicable only for the Adoptive networks only, because in adoptive network able to change the weight but is not possible in fixed networks (her weight is fixed). As per the adoptive network the learning process can be divided into two type namely supervised learning & unsupervised learning. Priority is going to the supervised learning because is a offline but unsupervised is performed in online.

## E. Transfer Function

Transfer function is used to identify the behavior of a Neural Network which is normally used in between hidden layer to the output unit or layer. This function falls in one of the following three ways namely, linear unit, threshold, sigmoid.

## F. Back propagation algorithm

Training is required for the neural network to perform some task. So the back propagation algorithm is the best one technique or algorithm we must adjust the weight of each unit in such a way that the error between the desired output and actual output is reduced.

## V. PARTICLE SWARM OPTIMIZATION

Particle Swarm Optimization (PSO) is a Meta heuristic technique originally designed to solve non-linear continuous optimization problems which is introduced by Kennedy & Eberhart. Later on PSO is used as swarm movement and intelligence purpose [6-8]. The terms which are related to PSO are Position, Velocity, Multi dimensional space and Operator etc. In PSO basic operations are:

- Neighborhood particle association with each particle.
- Each particle knows the fitness of neighborhood particle.
- And the best fitness function is used to find the velocity of particles.

## A. PSO Algorithm

**Step-1:** Initialize the population size (Number of particle)
**Step-2:** Evaluate the objective function
**Step-3:** Initialize the velocity of particle.
**Step-4:** Find the *Pbest & Gbest*.
**Step-5:** Find the new position of particle and velocity
**Step-6:** Repeat step-2 to step-5 till iteration ends or condition satisfied.

## B. Basic Flow

**Step-1:** Initialize the swarm by randomly assigning initial velocity and a position to each particle in each dimension.
**Step-2:** Evaluate the desired fitness function for each particle's position to be optimized.
**Step-3:** For each individual particle, update its historically best position $P_i$, if its current position is better than its historically best one.
**Step-4:** Update the swarm's global best particle, and reset its index as $g$ and its position at $p_g$.
**Step-5:** Update the velocities of particles.
**Step-6:** Update each particle position.
**Step-7:** Repeat step 2-6 until stopping criteria satisfied.

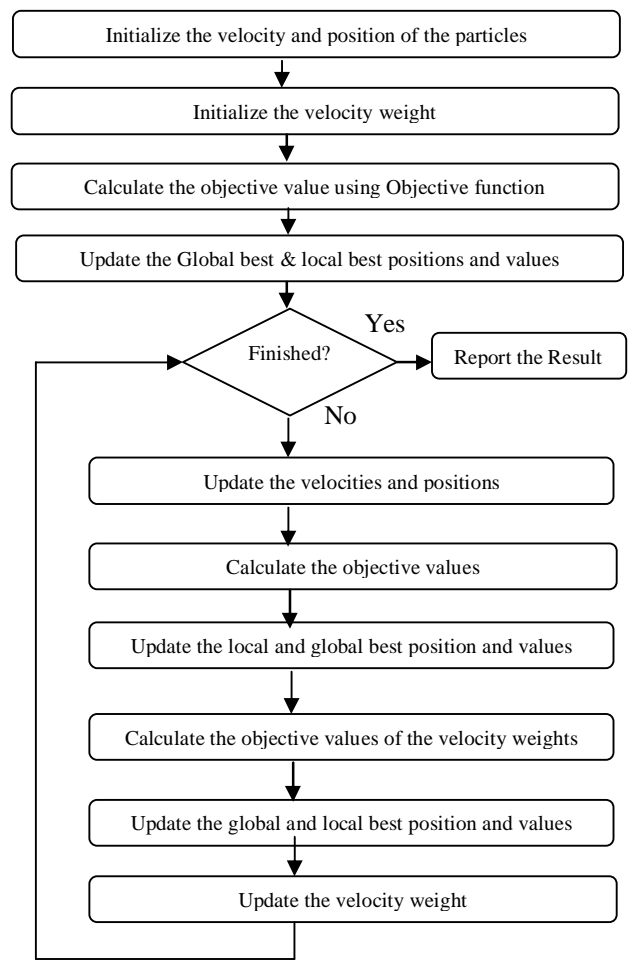

Fig.3 Flow Chart of PSO

- **Particle:** Particle is the size of the population that we have considered for our research work. Here each particle we considered as NN. And the particle position can be calculated by using the following updated formulae:

$$x_i^{t+1} = x_i^t + v_i^{t+1}$$

Where, $x_i^t$ is the current position of particle.
$v_i^{t+1}$ is velocity of particle

- **Local or Personal Best ($P_{best}$):** This is the each individual best of each particle that we have considered.
- **Global Best ($G_{best}$):** Global best is the minimum or maximum among all the particles according to the local best value of particles.
- **Velocity:** Velocity is calculated for finding the new position of particle [9].

$$v_i^{t+1} = v_i^t + c_1 r_1^t [P_{best,i}^t - x_i^t] + c_2 r_2^t [G_{best}^t - x_i^t]$$

Where,
  $i$ is particle index to identify particle
  $t$ is Iteration Number
  $C_1$ and $C_2$ are acceleration constant for cognitive & social component. And range generally between [0-2]. And considered values are 1.4962 or 1.494
  $r_1$ and $r_2$ are the random values range between [0-1]

- **Objective/Fitness function:** Fitness function helps us to periodically update the personal best and global best of the particle. Fitness functions are used as a performance test problem for optimization algorithm (preferably, in NN). The fitness functions are generally in non-negative.

## VI. PROPOSED SOLUTION

As we are working in the area of financial forecasting by using soft computing techniques.. So we have to select best one soft computing technique to find the forecasted result one. So following are the steps for our proposed model:

- Initialize each particle by a uniform NN
- Use PSO for finding of Global best one particle and that global best particle one becomes the best one NN.
- By using this best one NN we forecasted our BSE_SENSEX data.
- For our proposed method or solution we have considered two algorithm namely Initialization of NN to each particle & finding Global best NN for forecasting.

The algorithm we named as PSO-Gbest-NN, where we are able to forecast or predict the future outcomes of BSE_SENSEX (we have considered). From Jan 15 to Apr 15.

**Algorithm-1** (*Initialize*)

1: for each particle $i$ in $S$ do
2:     for each dimension $d$ in $D$ do
3:         $x_i = r(x_{min}, x_{max})$
4:         $v_i = r(-v_{max}/3, v_{max}/3)$
5:     end for
6:     $P_{best,i} = x_i$
7:     if $f(P_{best,i}) < f(G_{best})$ then
8:         $G_{best} = P_{best,i}$
9:     end if
10: end for

**Algorithm-2** (*PSO-Gbest-NN*)

1: Initialize
2: repeat
3:     for each particle $i$ in $S$ do
4:         if $f(x_i) < f(P_{best,i})$ then   //update Pbest position
5:             $P_{best,i} = x_i$
6:         end if
7:         if $f(P_{best,i}) < f(G_{best})$ then //update the Gbest position
8:             $G_{best} = P_{best,i}$
9:         end if
10:    end for
11:    for each particle $i$ in $S$ do // update velocity
12:        for each dimension $d$ in $D$ do // update positio
13:            $v_i^{t+1} = v_i^t + c_1 r_1^t [P_{best,i}^t - x_i^t] + c_2 r_2^t [G_{best}^t - x_i^t]$
14:            $x_i^{t+1} = x_i^t + v_i^{t+1}$
15:        end for
16:    end for
17:    $t = t+1$
18: until $t < MAX\_ITERATIONS$

## VII. EXPERIMENTAL RESULT

After successfully applied the proposed algorithm to the BSE_SENSEX data we are able to get following table with deduction of error by nearly 79%.

| BSE_SENSEX DATA_MONTHLY | | | | |
|---|---|---|---|---|
| Month | Open | High | Low | Close |
| Jun11 | 19859.22 | 19860.19 | 18467.16 | 19395.81 |
| Jul11 | 19352.48 | 20351.06 | 19126.82 | 19345.7 |
| Aug11 | 19443.29 | 19569.2 | 17448.71 | 18619.72 |
| Sept11 | 18691.83 | 20739.69 | 18166.17 | 19379.77 |
| Oct11 | 19452.05 | 21205.44 | 19264.72 | 21164.52 |
| Nov11 | 21158.81 | 21321.53 | 20137.67 | 20791.93 |
| . | | | | |
| Dec14 | 20771.27 | 21483.74 | 20568.7 | 21170.68 |
| **Jan 15** | **20390.19** | **21379.63** | **20358.84** | **21058.72** |
| **Feb 15** | **20191.67** | **21314.99** | **20213.70** | **20979.06** |
| **Mar 15** | **20088.54** | **21275.23** | **20113.81** | **20922.90** |
| **Apr 15** | **20034.95** | **21250.78** | **20045.08** | **20883.30** |

Table 2 Forecasted values

In this above table the forecasted result after using our proposed algorithm is mentioned by using bold colored one. From Jan 15 to Apr 15 for the entire column namely Open, High, Low and Close. As we have seen our proposed algorithm has been worked perfectly.

| BSE_SENSEX DATA MONTHLY | | |
|---|---|---|
| Month | OPEN with Error(.4369) | OPEN with Error(.3480) | OPEN reduced Error |
| Jan 15 | 20390.62 | 20390.53 | **20390.19** |
| Feb 15 | 20192.10 | 20192.01 | **20191.67** |
| Mar 15 | 20088.97 | 20088.88 | **20088.54** |
| Apr 15 | 20035.38 | 20035.29 | **20034.95** |

Table 3 OPEN SENSEX Forecast values

| BSE_SENSEX DATA MONTHLY | | |
|---|---|---|
| Month | HIGH with Error(.4369) | HIGH with Error(.3480) | HIGH reduced Error |
| Jan 15 | 21380.06 | 21379.97 | **21379.63** |
| Feb 15 | 21315.42 | 21314.34 | **21314.99** |
| Mar 15 | 21275.66 | 21275.57 | **21275.23** |
| Apr 15 | 21251.21 | 21251.12 | **21250.78** |

Table 4 HIGH SENSEX Forecast values

| BSE_SENSEX DATA MONTHLY | | |
|---|---|---|
| Month | LOW with Error(.4369) | LOW with Error(.3480) | LOW reduced Error |
| Jan 15 | 20359.27 | 20359.18 | **20358.84** |
| Feb 15 | 20214.13 | 20214.04 | **20213.70** |
| Mar 15 | 20114.24 | 20114.15 | **20113.81** |
| Apr 15 | 20045.16 | 20045.42 | **20045.08** |

Table 5 LOW SENSEX Forecast values

| BSE_SENSEX DATA MONTHLY | | |
|---|---|---|
| Month | CLOSE with Error(.4369) | CLOSE with Error(.3480) | CLOSE reduced Error |
| Jan 15 | 21059.15 | 21059.06 | **21058.72** |
| Feb 15 | 20979.49 | 20979.40 | **20979.06** |
| Mar 15 | 20923.33 | 20923.24 | **20922.90** |
| Apr 15 | 20883.73 | 20883.64 | **20883.30** |

Table 6 CLOSE SENSEX Forecast values

## VIII. CONCLUSION AND FUTURE WORK

As we have studied that the forecasting of BSE_SENSEX data we have predicted by using the proposed algorithm PSO-Gbest-NN is nearly accurate, that we have mentioned in this above tabulation with bold latter one. But our future work that we will do on this work with some advance on PSO or some other optimization technique like Evolutionary Algorithm, Differential Evolutionary Algorithm so that the error like 0.4369 or 0.3480 will be minimize or we will try to make these error to be zero.

AUTHORS PROFILE

**S. Gopal Krishna Patro** is a M.Tech scholar in Department of CSE & IT, VSSUT, Burla with specialization ICT. His area of of interest is Financial Forecasting, Machine Learning, Cloud Computing. He did his B. Tech. in CSE from RIT Berhampur and Diploma in CSE from UCPES. He is having total 2 year of experience including both industry and teaching.

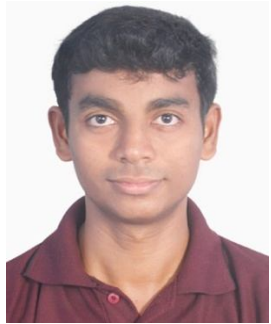

**Pragyan Parimita Sahoo** is a M.Tech scholar in department of CSE & IT, VSSUT, Burla. Her areas of intrest are Operating System Artificial Intelligence, Financial Forecasting. She did her B.Tech in IT from REC, Bhubaneswar under BPUT.

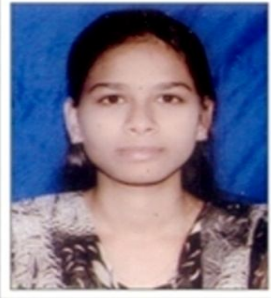

**Ipsita Panda** is also a M.Tech scholar in department of CSE & IT, VSSUT, Burla. Her areas of intrest are Database, Gesture Recognisation, Financial Forecasting, Artificial Intelligence. She did her B.Tech in IT from TTS, Bhubaneswar under BPUT.

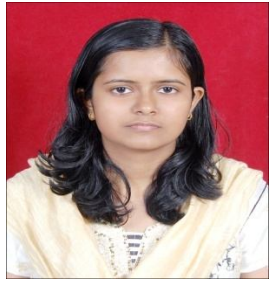

**Mr. Kishore Kumar Sahu** is a Assistant Professor in CSE & IT department, VSSUT, Burla. He is total 10 year of teaching experience at UG & PG level. He is persuing his Ph.D in Computer Science & Engineering Department. His areas of intrests are Soft Computing, Artificial Inteligence, Compiler & Theory of Computation.

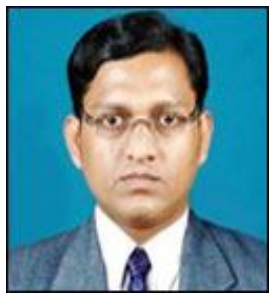